\documentclass{article}

\usepackage[preprint]{neurips_2023}

\usepackage[utf8]{inputenc} 
\usepackage[T1]{fontenc}    
\usepackage{url}            
\usepackage{booktabs}       
\usepackage{amsfonts}       
\usepackage{nicefrac}       
\usepackage{microtype}      
\usepackage{xcolor}         
\usepackage{graphicx}
\usepackage{subcaption}
\usepackage{algorithm, algorithmicx, algpseudocode}
\usepackage{makecell}
\usepackage{mathtools}

\usepackage{amsmath,amssymb,amsthm,amsfonts,mathrsfs,bm,multirow}

\DeclareMathOperator*{\minimize}{\text{minimize}}

\usepackage{url}

\usepackage[colorlinks=true,linkcolor=blue, citecolor=blue]{hyperref}

\title{SparseOptimizer: Sparsify Language Models through Moreau-Yosida Regularization and Accelerate via Compiler Co-design}

\author{%
  Fu-Ming Guo\thanks{For prompt response, please forward your email to fuming.guo.acad@gmail.com} \\
  Harvard University\\
  Lotus AGI\\
  \texttt{fug626@g.harvard.edu} \\
}

\begin{document}

\maketitle

\begin{abstract}
  This paper introduces SparseOptimizer, a novel deep learning optimizer that exploits Moreau-Yosida regularization to naturally induce sparsity in large language models such as BERT, ALBERT and GPT. Key to the design of SparseOptimizer is an embedded shrinkage operator, which imparts sparsity directly within the optimization process. This operator, backed by a sound theoretical framework, includes an analytical solution, thereby reinforcing the optimizer's robustness and efficacy. Crucially, SparseOptimizer's plug-and-play functionality eradicates the need for code modifications, making it a universally adaptable tool for a wide array of large language models. Empirical evaluations on benchmark datasets such as GLUE, RACE, SQuAD1, and SQuAD2 confirm that SparseBERT and SparseALBERT, when sparsified using SparseOptimizer, achieve performance comparable to their dense counterparts, BERT and ALBERT, while significantly reducing their parameter count. Further, this work proposes an innovative optimizer-compiler co-design strategy, demonstrating the potential of inference acceleration (\textbf{3.37x}, \textbf{6.30x}, and \textbf{7.15x} in comparison with Pytorch, TensorFlow, and LLVM generic compile, respectively) in SparseBERT when paired with an appropriately designed compiler. This study represents a significant step forward in the evolution of efficient, scalable, and high-performing large language models, setting a precedent for future exploration and optimization in this domain. The SparseOptimizer code and SparseALBERT model will be publicly available upon paper acceptance.
\end{abstract}

\begin{figure}[ht]
    \centering
    \includegraphics[width=13.5cm]{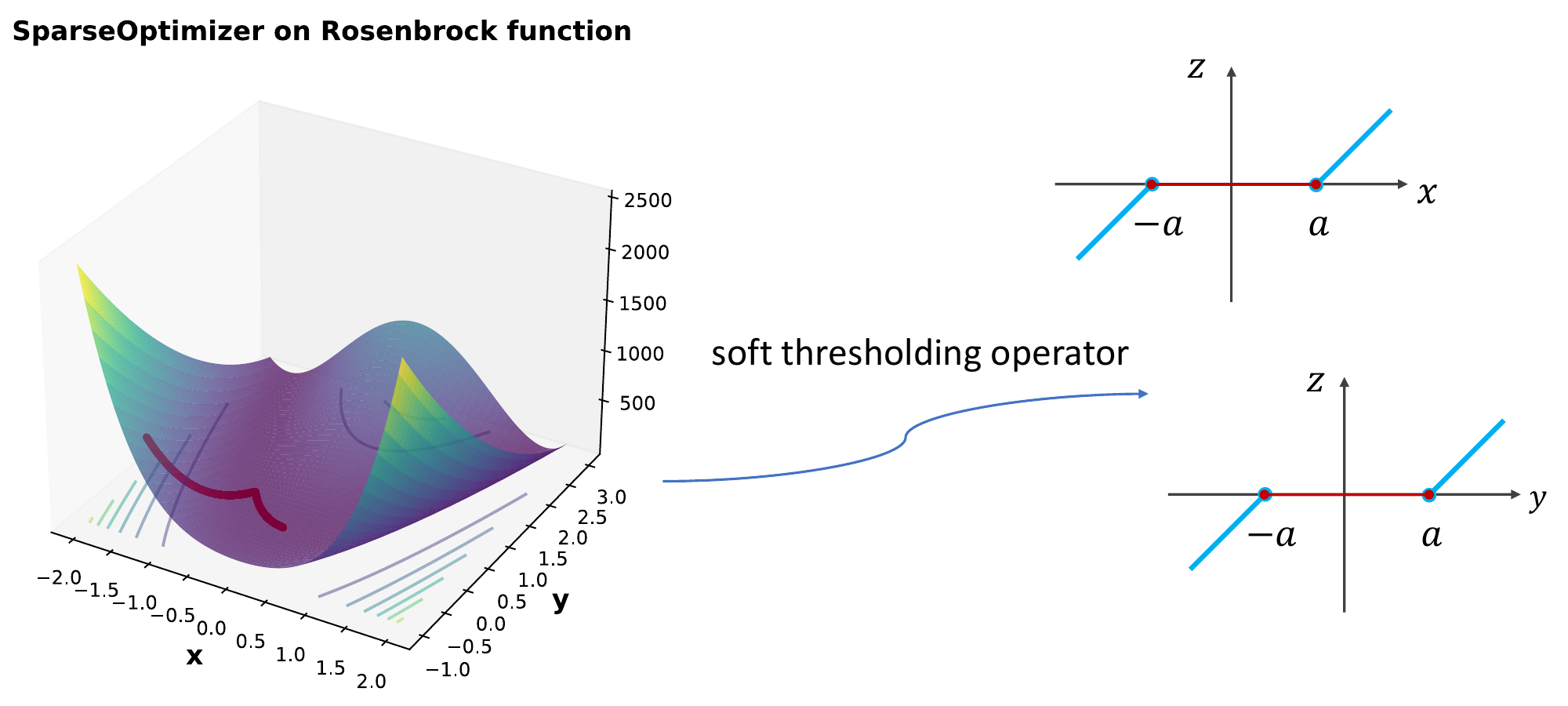}
    \caption{Overview of optimization process using SparseOptimizer on Rosenbrock function \citep{rosenbrock1960automatic}. A soft thresholding operator is incorporated to induce enhanced (either irregular or block) sparsity. This is an simplified illustration under 3D, compared to our application in optimizing language models. About the derivation and specific value choosing for $a$, please refer Section \ref{method}.}
    \label{fig:rwone_sparse_optimizer_overview}
\end{figure}
\section{Introduction}
Large Language Models (LLMs) such as BERT \citep{devlin2018bert}, ALBERT  \citep{lan2019albert}, GPT-3  \citep{brown2020language}, LaMDA  \citep{thoppilan2022lamda}, ChatGPT \citep{ouyang2022training}, PaLM2 \citep{anil2023palm} and GPT-4 \citep{gpt4} have emerged as key tools in a multitude of applications, ranging from text generation and translation to more complex tasks like summarization and conversation. These models are typically based on the Transformer architecture \citep{bahdanau2014neural, vaswani2017attention}, which leverages attention mechanisms to capture long-range dependencies in sequences. Despite their powerful capabilities, the deployment of these LLMs is often hindered by their substantial computational requirements and the substantial number of parameters, raising challenges in terms of memory footprint, computational efficiency, economic and environmental cost \citep{patterson2021carbon}.

To address these challenges, numerous methods, including pruning \citep{han2015deep, franklelottery, guo2019reweighted, guo2021algorithm}, distillation \citep{hinton2015distilling}, and quantization \citep{kim2021bert, dettmers2023qlora}, have been proposed to reduce model complexity without substantial loss of performance. The lottery ticket hypothesis was proposed by \citep{frankle2018lottery, frankle2020linear}, which observes that a subnetwork of randomly-initialized network can replace the original network with the same performance. \cite{chen2020lottery, chen2021lottery} demonstrate the core LTH observations remain generally relevent in transformer models for both computer vision and natural language processing. Pruning aims to reduce model size by removing less important connections or neurons, while distillation transfers knowledge from a larger model (teacher) to a smaller one (student). Quantization, on the other hand, reduces the precision of the model's weights.
Although these methods have proven to be effective, a unified approach that seamlessly integrates model optimization with efficient execution remains elusive.

In light of this, we propose SparseOptimizer, a novel deep learning optimizer tailored for large language models. Drawing upon the mathematical properties of Moreau-Yosida regularization \citep{moreau1966fonctionnelles, parikh2014proximal,bacho2023generalization}, SparseOptimizer embeds a shrinkage operator within the optimization process. This design choice naturally introduces enhanced sparsity into the model parameters, reducing complexity and computational demand, without necessitating any modification to the original model code. Therefore, the direct plug-and-play nature of SparseOptimizer broadens its applicability to a wide range of large language models.

The theoretical soundness of SparseOptimizer lies in its foundation on Moreau-Yosida regularization. By integrating proximal gradient methods \citep{martinet1970breve, boyd2004convex, parikh2014proximal} methods with weighted $\ell_1$ minimization \citep{candes2008enhancing} within this regularization framework, our formulation yields a unique analytical solution manifesting as a shrinkage operator \citep{boyd2011distributed}. This offers not only an intuitive understanding of the optimizer's behavior but also a solid mathematical grounding that guarantees the convergence properties of the optimization process. The incorporation of the shrinkage operator seamlessly introduces enhanced sparsity into the model during the optimization, laying the groundwork for model pruning and efficiency.

In a unique approach towards efficient model deployment, we advocate for an optimizer-compiler co-design strategy. We have engineered a compiler specifically designed to leverage the sparsity in models optimized with SparseOptimizer, thereby enabling further acceleration in inference time. This is exemplified in our empirical studies on SparseBERT Base and SparseBERT Large, where the harmonization between SparseOptimizer and our compiler design allows for remarkable inference acceleration.

This paper's main contributions can be summarized as follows:
\begin{itemize}
    \item We propose SparseOptimizer, a novel deep learning optimizer designed specifically for LLMs. The optimizer integrates Moreau-Yosida regularization into the optimization process, naturally inducing sparsity in the model parameters.
    \item The SparseOptimizer's plug-and-play design makes it a directly pluggable tool for LLMs without any modification needed to the original model code. This significantly broadens the applicability of our approach to a wide range of large language models.
    \item We pioneer an optimizer-compiler co-design strategy for efficient execution of sparse models, which is further exemplified through empirical evaluations on SparseBert Base and SparseBert Large.

\end{itemize}

This work represents a significant stride in the pursuit of practical and efficient deployment of large language models, shedding light on new directions for research and development in this realm.

\section{Related Work}
\label{gen_inst}

\subsection{Distillation }
Knowledge distillation, a concept first introduced by \cite{hinton2015distilling}, has emerged as an effective model compression technique whereby a smaller student model is trained to mimic the functionality of a larger, more complex teacher model. Subsequent research has expanded upon this original premise, with distillation methodologies now typically falling into two main categories: general distillation \citep{sanh2019distilbert, sun2020mobilebert, wang2020minilm} and task-specific distillation \citep{jiao2019tinybert}.

General distillation leverages unlabeled data as a vehicle for knowledge transfer. Despite the computational expense, these methods have shown to be essential in retaining performance levels, particularly when pre-training the student network on an unlabeled corpus \citep{turc2019well, jiao2019tinybert}. On the other hand, task-specific distillation exploits task-relevant data to instill the knowledge from the teacher model to the student model.

However, the real potential of knowledge distillation has begun to shine through in recent research that combines these two distinct approaches. For instance, the work of \cite{jiao2019tinybert} showed that such hybrid methods could push the performance boundaries even further.

The advancements in distillation techniques have paved the way for increasingly efficient utilization of large language models. One of the notable milestones in this evolution is the study by \cite{hsieh2023distilling}, which demonstrated that superior performance could be achieved even with less training data and smaller model sizes. 

In summary, the continual progress in knowledge distillation research has opened up new avenues for large language model efficiency. It is this body of work that serves as a valuable foundation for our novel SparseOptimizer, offering a new perspective on model sparsification and optimization.

\subsection{Pruning}

Pruning is a well-established technique for reducing the computational complexity of machine learning models. This method, initially developed for neural networks \citep{lecun1989optimal, hassibi1992second}, has recently been extended to transformer-based models \citep{franklelottery, chen2020lottery, guo2021algorithm}. A variety of methods have been proposed, ranging from structured pruning that removes whole layers or attention heads \citep{michel2019sixteen, voita2019analyzing}, to unstructured pruning that removes individual weights \citep{zhu2017prune}. These approaches offer a trade-off between the reduction in computational complexity and the ability to retain performance.

The SparseOptimizer we propose in this paper differs significantly from these pruning techniques. While traditional pruning methods are often performed post-training, requiring an additional fine-tuning stage to recover model performance, SparseOptimizer integrates sparsity induction directly into the optimizer and can be done in the training process. This is achieved by embedding a shrinkage operator within the optimizer, which is motivated by Moreau-Yosida regularization and naturally induces sparsity during training. Such an approach has the advantage of eliminating the need for a separate pruning and fine-tuning process, thus simplifying the overall model optimization pipeline.

Additionally, the plug-and-play nature of SparseOptimizer, which allows it to be applied directly at the optimization level, provides a level of versatility absent in typical pruning methods. This feature enables SparseOptimizer to be readily applicable across a wide range of large language models without necessitating any specific model architecture changes or any particular coding of the model.

Finally, while pruning techniques often yield sparse models that require specialized hardware or software to fully exploit the sparsity \citep{gale2020sparse, guo2021algorithm}, we propose an optimizer-compiler co-design approach. We design a compiler specifically to take advantage of the sparsity induced by our optimizer, providing an additional layer of optimization and resulting in a significant inference acceleration in SparseALBERT, our sparsified version of ALBERT.

In summary, SparseOptimizer provides a unique, effective, and broadly applicable method for reducing the parameter footprint of large language models, offering a compelling alternative to existing pruning approaches.

\section{SparseOptimizer}
\label{method}

Following the pioneering pruning work \citep{lecun1989optimal, hassibi1992second, strom1997phoneme}, and recent advancements \citep{han2015deep, franklelottery}, we formulate our SparseALBERT target as the subsequent optimization problem:
\begin{align}\label{eq: prob_EoT}
    \begin{array}{ll}
\displaystyle \minimize_{\mathbf{w} }        &  f( \mathbf{w}; \mathcal{D}) + \mu h ( \mathbf{w}),
    \end{array}
\end{align}

where $\mathbf{w}$ is the weights of the Deep Neural Network,  $f_i( \mathbf{w}; \mathcal{D})$ is the loss (can be either pretraining loss or finetuning loss) function, and $h(\mathbf{w})$ is the introduced sparsity-promoting penalty function. $\mu$ is a balancing parameter, and the weights $\mathbf{w}$ becomes sparser as $\mu$ increases over $[0, +\infty)$. Typically in the field of neural network pruning, $h(\mathbf{w})$ is the variances of $\ell_1$ norm or $\ell_2$ norm. 

Here we choose a weighted $\ell_1$ norm of weights $\mathbf{w}$ as the sparsity-promoting penalty function, as
\begin{align}
    \label{eq: ell1_introduce}
    h(\mathbf{w}) = \displaystyle \sum_{} \gamma^\ell \| \mathbf{w} \|_{\ell_{1}}
\end{align}
and $  \gamma^\ell \|\mathbf{w} \|_{\ell_{1}}$ is the weighted $\ell_1$ norm of the weights $\mathbf{w}$, $\gamma^\ell$ is the weighting parameter for the weighted $\ell_1$ norm, different from the weights $\mathbf{w}$. For the weighting factor $\gamma^{\ell}$ in the weighted norm $  \gamma^\ell \|\mathbf{w} \|_{\ell_{1}}$, the previous form is for simplification, and the concrete form is defined in an iteratively progressing manner \citep{candes2008enhancing}: for each single entry $w_i$ in the matrices of weights $\mathbf{w}^{\ell}$, there is a corresponding weighting factor $\gamma_i^{\ell+1}$, and
\begin{align}
\gamma_i^{\ell+1}=\frac{1}{|w_i^{(\ell)}| + \epsilon},
\end{align}
here $\ell$ is the iteration counting integer starting from 0, and $\epsilon$ is a small constant, e.g., $\epsilon = 0.0001$. The iterative process will terminate on convergence or when $\ell$ attains a specified maximum number of iterations $\ell_{max}$. 

We note that $f_i( \mathbf{w}; \mathcal{D})$ is a differentiable function, while $\gamma^\ell \| \mathbf{w} \|_{\ell_{1}}$ is a non-differentiable function due to the $\ell_1$ norm. Then we introduce the Moreau-Yosida regularization of $\gamma^\ell \| \mathcal{W} \|_{\ell_{1}}$ as
\begin{align}\label{eq: Moreau-Yosida} 
\begin{array}{ll}
\displaystyle \minimize_{\mathbf{w}, \mathbf{z} \in \mathbb{R} } &
     h_\lambda(\mathbf{w}) =   \gamma^{\ell} \|\mathbf{z}\|_{\ell_1} + \frac{1}{2\lambda}\|\mathbf{z}- \mathbf{w}\|^2.
\end{array}
\end{align}

Here, $\mathbf{z}$ is the auxiliary variable used in the Moreau-Yosida regularization, and plays a role as a stand-in for weights $\mathbf{w}$ during the minimization process. One of the key properties of the Moreau-Yosida regularization is that it is differentiable even when the original function is not. So the whole objective function is differentiate now. Thus, the equation \eqref{eq: prob_EoT} is equivalent to:
\begin{align}\label{eq:problem_re}
\begin{array}{ll}
\displaystyle \minimize_{\mathbf{w}, \mathbf{z} \in \mathbb{R} }        &  f( \mathbf{w} ; \mathcal{D}) + \mu \{ \gamma^{\ell} \|\mathbf{z}\|_{\ell_1} + \frac{1}{2\lambda}\|\mathbf{z}- \mathbf{w}\|^2\},
    \end{array}
\end{align}

Thus, the gradient of the Moreau-Yosida regularization can be computed as follows:
\begin{align}\label{eq:my_gradient}
\nabla h_{\lambda}(\mathbf{w}) = \frac{1}{\lambda_k} (\mathbf{w} - \text{prox}_{\lambda_k \| \gamma^\ell \cdot \|_{\ell_{1}}}(\mathbf{w})),
\end{align}
and the gradient update is rewritten as follows:
\begin{align}\label{eq:proximal_gradient}
\mathbf{w}^{k+1} = \mathbf{w}^k - \lambda_k (\nabla f(\mathbf{w}^k) + \nabla h_{\lambda}(\mathbf{w}^k)),
\end{align}
where the $\lambda(\lambda > 0)$ is the learning rate. For simplification and the adaptability to different optimizer in neural network training, we set $\lambda = \lambda_k$ in equation \eqref{eq:problem_re}.

Recall that in \eqref{eq: Moreau-Yosida}, we say $\mathbf{z}$ is the auxiliary variable and is a stand-in for weights $\mathbf{w}$ during the minimization process. Hence, after proximal projection
\begin{align}\label{eq: z_update}
\mathbf{z} = \mathbf{w}^k - \lambda (\nabla f(\mathbf{w}^k)).
\end{align}

In equation \eqref{eq: Moreau-Yosida}, since both $h(\mathbf{w})$ the the Frobenius norm can be written as a summation of component-wise functions of a matrix, we can decompose \eqref{eq: Moreau-Yosida} into sub-problems expressed in terms of the individual elements of $\mathbf{w}$. Thus  \eqref{eq:my_gradient} \eqref{eq:proximal_gradient} and \eqref{eq: z_update} can also be rewritten in terms of the individual elements $w_i$ of of $\mathbf{w}$.

The unique analytical solution to \eqref{eq:problem_re} is given by a \textcolor{blue}{soft thresholding operator},  and also a \textcolor{blue}{shrinkage operator}(e.g., see \cite{boyd2011distributed}, section 4.4.3, \cite{lin2013design}, eq 14, and \cite{liu2014sparsity}, eq 21):
\begin{equation} 
\label{eq:analytical_s}
w_{i}^{*} = \left\{ \begin{array} { l l } { \left( 1 - \frac {  \lambda \gamma _ { i } } { \mu\left| z _ { i } \right| } \right) z _ { i } } & { \left| z _ { i } \right| >  \frac{\lambda}{\mu} \gamma _ { i } } \\ { 0 } & { \left| z _ { i } \right| \leq \frac{\lambda}{\mu} \gamma _ { i } }. \end{array} \right.
\end{equation}

\textbf{Remark}  In \textbf{block sparse design}, $\mathbf{w}$ is determined by:

\begin{equation}
    \sum_{i, j} \gamma_{ij}^{\ell} \|w_{ij}\|_F
\end{equation}
and the minimizers of equation \eqref{eq: Moreau-Yosida} are obtained by replacing the absolute value of $z_i$ in equation \eqref{eq:analytical_s}, with the \textit{Frobenius norm $\| \cdot \|_F$ of the corresponding block submatrix $z_{ij}$}.
    
Finally, a \textcolor{blue}{soft thresholding operator},  and also a \textcolor{blue}{shrinkage operator} emerges, with our  formulation provides the theoretical foundation. The incorporation of this analytical solution into the AdamW optimizer \citep{loshchilov2017decoupled} used in BERT, ALBERT, GPT and many other LLMs, yields our updated optimizer Algorithm \ref{alg:proximal_operator}:

\begin{algorithm}
    \caption{SparseOptimizer}
    \label{alg:proximal_operator}
\begin{algorithmic}[1]
\State \textbf{Given} $\alpha=0.001, \beta_{1}=0.9, \beta_{2}=0.999, \epsilon = 10^{-6}, \lambda \in \mathbb{R}$
\State \textbf{Initialize} time step $k \leftarrow 0$, parameters of pre-trained model $\mathbf{w}$, first moment vector $\mathbf{m}_{t=0} \leftarrow \mathbf{0}$, second moment vector $\mathbf{v}_{t=0} \leftarrow \mathbf{0}$, schedule multiplier $\eta_{k=0} \in \mathbb{R}$
\Repeat
\State $k\leftarrow k+1$
\State $\nabla f _ { k } \left( \boldsymbol { \mathbf{w} } _ { k - 1 } \right) \leftarrow \text { SelectBatch } \left( \boldsymbol { \mathbf{w} } _ { k - 1 } \right)$
\State $\boldsymbol { g } _ { k } \leftarrow \nabla f _ { k } \left( \boldsymbol { \mathbf{w} } _ { k - 1 } \right)$
\State $\boldsymbol { m } _ { k } \leftarrow \beta _ { 1 } \boldsymbol { m } _ { k - 1 } + \left( 1 - \beta _ { 1 } \right) \boldsymbol { g } _ { k }$
\State $\boldsymbol { v } _ { k } \leftarrow \boldsymbol { \beta } _ { 2 } \boldsymbol { v } _ { k - 1 } + \left( 1 - \beta _ { 2 } \right) \boldsymbol { g } _ { k } ^ { 2 }$
\State $\hat { \boldsymbol { m } } _ { k } \leftarrow \boldsymbol { m } _ { k } / \left( 1 - \beta _ { 1 } ^ { k } \right)$
\State $\hat { \boldsymbol { v } } _ { k } \leftarrow \boldsymbol { v } _ { k } / \left( 1 - \beta _ { 2 } ^ { k } \right)$
\State $\eta_{k} \leftarrow \operatorname{SetScheduleMultiplier}(k)$
\State $\mathbf{z} \leftarrow \boldsymbol { \mathbf{w} } _ { k - 1 } - { \eta } _ { k } \left( \alpha \hat { \boldsymbol { m } } _ { k } / ( \sqrt { \hat { \boldsymbol { v } } _ { k } } + \epsilon ) + \lambda \boldsymbol { \mathbf{w} } _ { k - 1 } \right)$
\State $\mathbf{w}_{k}\leftarrow \operatorname{prox}_{\lambda \| \gamma^\ell \cdot \|_{\ell_{1}}} (\mathbf{z})$:
\State \hspace{\algorithmicindent}\hspace{\algorithmicindent}$w_{i}^{*} = \left\{ \begin{array} { l l } { \left( 1 - \frac {  \lambda \gamma _ { i } } { \mu\left| z _ { i } \right| } \right) z _ { i } } & { \left| z _ { i } \right| >  \frac{\lambda}{\mu} \gamma _ { i } } \\ { 0 } & { \left| z _ { i } \right| \leq \frac{\lambda}{\mu} \gamma _ { i } }. \end{array} \right.$
\State \hspace{\algorithmicindent}\hspace{\algorithmicindent}where $\gamma_i^{\ell+1}=\frac{1}{|w_i^{(\ell)}| + \epsilon}$, $\ell$ is the iteration counting integer starting from 0, \State \hspace{\algorithmicindent}\hspace{\algorithmicindent}and $\epsilon$ is a small constant, e.g., $\epsilon = 0.0001$
\Until stopping criterion is met
\State \Return optimized sparse model $\mathbf{w}$
\end{algorithmic}
\end{algorithm}

\section{Experiments}

\subsection{Promoting Sparsity by SparseOptimizer}
Additional information pertaining to our experimental design, including specifics regarding the utilized datasets, model architecture, and hyperparameters, will be elaborated in the forthcoming sections. These details aim to provide an in-depth understanding of the framework and settings that guided our empirical results. In the spirit of reproducibility and to facilitate further investigation, we have included the license to our codebase within our submission. Upon the acceptance of this paper, we pledge to publicly release our code as well as the model checkpoints.

\textbf{Data:} In pre-training, we use the same pre-training corpora as \cite{devlin2019bert}: BookCorpus ($800\mathrm{M}$ words) \citep{zhu2015aligning} and English Wikipedia ($2,500\mathrm{M}$ words). Based on the same corpora, we use the same preprocessing script\footnote{https://github.com/google-research/albert} to create the pre-training data. In fine-tuning, we report our results on the Stanford Question Answering Dataset (SQuAD) \citep{rajpurkar2016squad, rajpurkar2018know}, ReAding Comprehension from Examinations (RACE) \citep{lai2017race} and the General Language Understanding Evaluation (GLUE) benchmark \citep{wang2018glue}. We use two versions of SQuAD: V1.1 and V2.0 \citep{rajpurkar2016squad, rajpurkar2018know}. The GLUE is a collection of datasets/tasks for evaluating natural language understanding systems\footnote{The datasets/tasks are: CoLA \citep{warstadt2018neural}, Stanford Sentiment Treebank (SST) \citep{socher2013recursive}, Microsoft Research Paragraph Corpus (MRPC) \citep{dolan2005automatically}, Semantic Texual Similarity Benchmark (STS) \citep{agirre2007semeval}, Quora Question Pairs (QQP), Multi-Genre NLI (MNLI) \citep{williams2017broad}, Question NLI (QNLI) \citep{rajpurkar2016squad}, Recognizing Textual Entailment (RTE) and Winograd NLI(WNLI) \citep{levesque2012winograd}.}.

\textbf{Input/Output representations:} Consistent with the representation setting established by \cite{devlin2018bert}, our methodology adopts identical strategies for both pre-training and fine-tuning stages. We employ WordPiece embeddings \citep{wu2016google} to tokenize our inputs, capitalizing on their ability to decompose language into manageable and expressive units. This approach operates with a vocabulary size of 30,000 tokens, striking a balance between the model's complexity and its ability to capture a broad spectrum of linguistic features.

We observe critical nuances in the implementation of WordPiece embeddings, particularly their utilization of special tokens. The [CLS] token invariably marks the beginning of each sentence, serving as an aggregation point for sentence-level information. Conversely, the [SEP] token acts as a divider between sentences, offering crucial contextual separation. Extending the parameterization strategy delineated by ALBERT \citep{lan2019albert}, we adopt a factorized embedding design. This approach decouples the dimensions of the hidden layers and the embedding layers, effectively reducing the overall parameter count while maintaining expressivity. Therefore, our representation of language, while efficient, remains potent and adaptable across a wide array of tasks.

\textbf{Evaluation} : For the pre-training stage, our evaluation methodology embraces two objective functions: Masked Language Modeling (MLM) and Sentence Order Prediction (SOP), following the ALBERT architecture. During MLM, a random subset of input sequence tokens is selected and substituted with the special $([\text{MASK}])$ token. The objective of MLM is to predict these masked tokens, assessed by a cross-entropy loss. The SOP task encourages the model to understand discourse-level coherence by comparing consecutive segments from the same document (positive examples) and the same segments with swapped order (negative examples).

\begin{table}[t]
    \centering
    \resizebox{\linewidth}{!}{
    \begin{tabular}{*{10}{c}}
    \toprule
    Model & Parameters & SQuAD v1.1 & SQuAD v2.0 & RACE & {Parameter Sharing} & \\
    \midrule
    BERT Base & 108M & 88.5/80.8 & 80.4/77.6  & - &  False \\ 
    BERT Large & 334M & 90.9/84.1 & 81.9/78.7 & - &  False \\ 
    ALBERT Base   & 12M & 89.3/82.3 & 80.0/77.1 & 64.0 & True \\
    ALBERT Large   & 18M & 90.6/83.9 & 82.3/79.4 & 67.5  & True\\
    CoFi Pruning    & 26M & 82.6/- & - & - & False\\
    Movement Pruning    & 8.5M & 79.9/69.5 & - & -  & False \\
    \midrule
    
    \textbf{SparseBERT} Large & 39M & 79.3 & 81.4 & - & True\\
    \textbf{SparseALBERT} Base $(t=4)$ & 6M & 87.4/79.1 & 80.0/77.1 & 66.6 & True \\
    \textbf{SparseALBERT} Base $(t=6)$ & 5M & 87.5/79.3 & 79.7/76.8 & 65.8 & True \\
      \textbf{SparseALBERT} Large & 8M & 88.0/80.0 & 79.5/76.5 & 66.7 & True \\
    \bottomrule
    \\
    \end{tabular}
    }
    \caption{SparseALBERT Base and SparseALBERT Large evaluation results on QA.}
    \label{tab:qa}
\end{table}

During pre-training, we employ MLM and SOP as the guiding objectives for training and evaluating the ALBERT model. Furthermore, in the fine-tuning stage, we utilize different evaluation metrics tailored for specific tasks. For instance, F1 scores gauge the performance on SQuAD, QQP, and MRPC, while Matthew's Correlation and Pearson-Spearman Correlation are utilized for CoLA and SST2 respectively. For the remaining tasks, accuracy scores serve as the evaluation metric.

\begin{table}[ht]{
    \centering
    {
    \resizebox{\linewidth}{!}{
    \begin{tabular}{*{10}{c}}
    \toprule
    Model & Parameters & CoLA & MNLI & MRPC & QNLI & QQP & RTE & SST-2 & STS-B  \\
    \midrule
    BERT Base & 108M & 52.1 & 84.6  & 86.7 & 90.5 & 71.2 & 66.4 & 93.5 & 85.8\\ 
    BERT Large & 334M & 62.5 & 86.7 & 87.8 & 92.7 & 72.1 & 70.1 & 94.9 & 86.5\\ 
    ALBERT Base   & 12M & - & 81.6 & - & - & - & - & 90.3 & - \\
    ALBERT Large   & 18M & 71.4 & 83.5 & 90.2 & 95.2 & 92.0 & - & 91.5 & -  \\
    CoFi Pruning    & 26M & 35.6 & 80.6 & 82.6 & 86.1 & 90.1 & 64.7 & 90.6 & 83.1 \\
    Movement Pruning    & 8.5M & - & 81.2 & - & - & 90.2 & - & - & - \\
    \midrule
    
    \textbf{SparseBERT} Large & 39M & 79.3 & 81.4 & 81.9 & 88.0 & 89.2 & 67.5 & 90.5 & \\
    \textbf{SparseALBERT} Base & 5M & 77.9 & 81.6 & 85.1 & 90.9 & 90.1 & 72.2 & 92.4 & 90.4 \\
      \textbf{SparseALBERT} Large & 8M & 78.4 & 83.7 & 86.3 & 91.2 & 90.4 & 74.0 & 90.8 & 90.4\\
    \bottomrule
    \\
    \end{tabular}
    }
    
    }
    \caption{SparseALBERT Base and SparseALBERT Large evaluation results on GLUE.}
    \label{tab:glue}
}
\end{table}

\textbf{Results: } SparseBERT and SparseALBERT models have demonstrated notable efficacy and efficiency across various NLP benchmarks, as exhibited in Tables \ref{tab:qa} and \ref{tab:glue}. Both models offer a compelling balance between performance and model size, thus representing excellent choices for resource-constrained environments.

In Table \ref{tab:qa}, the SparseBERT Large model exhibits competitive performance on SQuAD v1.1 and SQuAD v2.0 with only 39M parameters. This is a significant reduction in model size, around 8.5x less compared to BERT Large, showcasing a promising increase in model efficiency.

Furthermore, SparseALBERT models display excellent results on the SQuAD benchmarks with a significant reduction in parameters. The SparseALBERT Base model with a time parameter $t=4$ and $t=6$, and the SparseALBERT Large model consistently displayed competitive performances.

Notably, SparseALBERT models also excelled on the RACE dataset. The SparseALBERT Base models, with only 6M and 5M parameters for $t=4$ and $t=6$ respectively, achieved scores of 66.6 and 65.8,  outperforming the ALBERT Base model. The SparseALBERT Large model also accomplished a similar feat, with a score of 66.7, which is competitive its dense counterpart ALBERT Large.

Table \ref{tab:glue} highlights the performance of SparseBERT and SparseALBERT models on the GLUE benchmark. Both SparseBERT Large and SparseALBERT models achieved similar or even superior performance compared to several dense models such as ALBERT Base, and sparse models such as CoFi Pruning\citep{xia2022structured} and Movement Pruning\citep{sanh2020movement}, with significantly fewer parameters. In some GLUE tasks, SparseALBERT Large even outperformed the BERT Base model, further demonstrating the potency of our sparsity-induced models.

In summary, the results underscore the significant advantages of SparseBERT and SparseALBERT, specifically in terms of delivering competitive performance while significantly reducing the model size. This presents an attractive balance for applications where computational resources are limited, yet demanding a high model performance.

\subsection{Acceleration by Optimizer-Compiler Co-Design}

\begin{figure}[t]
    \centering
    \includegraphics[width=13.5cm]{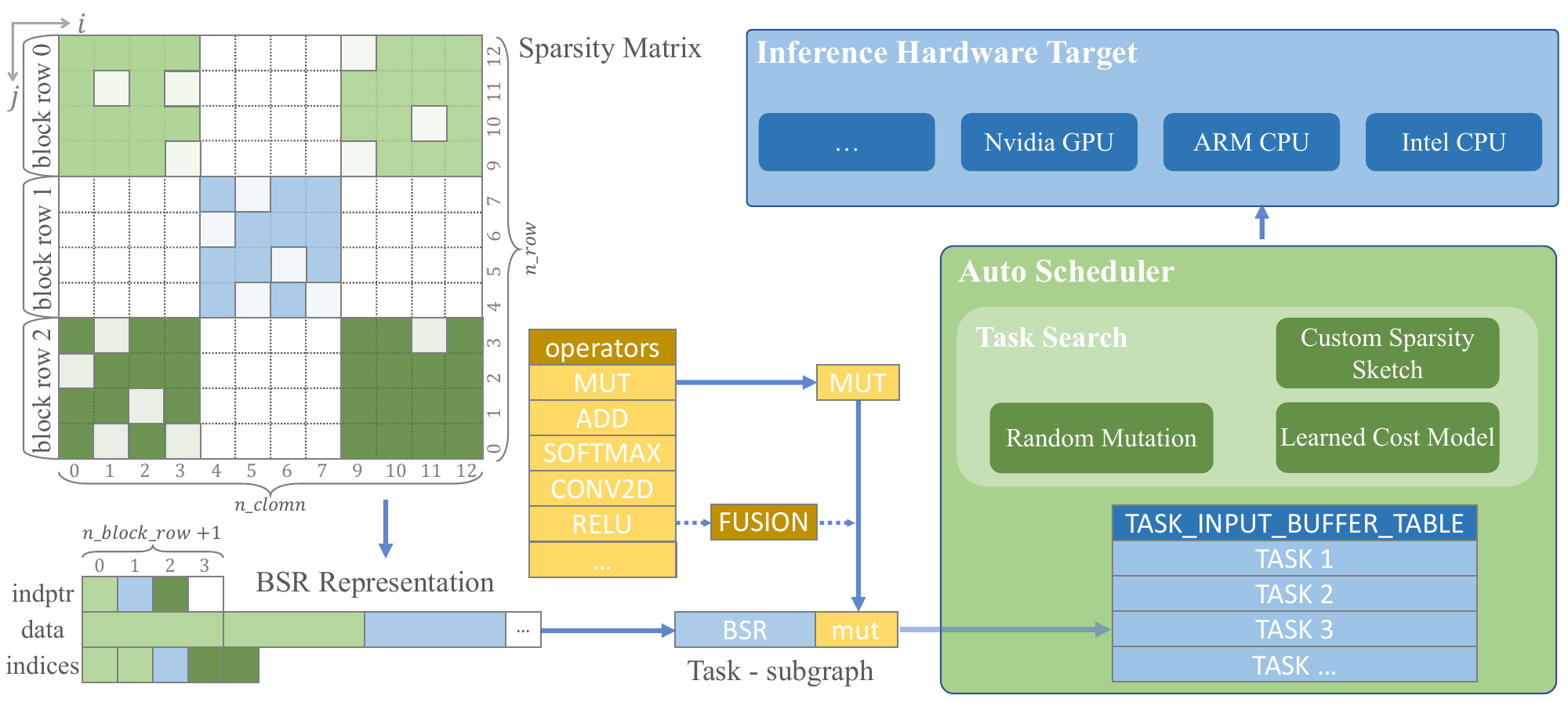}
    \caption{Overview of the augmented compiler: algorithm to compilation co-design}
    \label{fig:overview}
\end{figure}
\subsubsection{Optimizer-Compiler Co-Design Overview}

 Leveraging the principles of block sparsity in Transformer model execution on GPU, as delineated by \cite{gray2017gpu} and \cite{gale2020sparse}, combined with the empirical evidence demonstrating the efficacy of compiler scheduling in neural network inference acceleration by \cite{zhang2021full} and \cite{gale2020sparse}, we introduce enhancements to the TVM compiler \citep{chen2018tvm} to expedite inference in sparse neural networks. An overview of the compiler co-design architecture is in Figure \ref{fig:overview}. These enhancements include:
\begin{itemize}
    \item In our work, we have extended the functionality of Block Sparse Row (BSR) for its application with attention kernels and fully connected layers. The utility of BSR lies in its ability to shrink the memory footprint of sparse neural networks while enhancing inference speed. The key to sparse neural networks' acceleration involves the omission of operations such as element-wise matrix multiplication on pruned, zero-weight elements and utilizing sparsity structure-based operations.

    \item The TVM task scheduler is able to reuse structure-based sparsity. The aforementioned indices and indptr of BSR representation intrinsically reflect the characteristics of sparse matrices. The BSR representations are stored in a task buffer together with corresponding operators in TVM. TVM analyzes the similarity of tasks in the buffer and optimize the execution of the tasks through an auto-scheduler. The analysis proceeds in the task searching stage, attending to different hardware specifications (e.g., number of cores, cache size, instruction set architecture (ISA), max memory per block, and max thread per block). If two tasks in the task buffer are the same, TVM treats them as identical and reuse them. If two tasks are similar, TVM schedules them adjacent in the execution path.

    \item We utilize LLVM and Relay IR to compile the computation graph, focusing on generic CPU architectures as well as those supporting AXV2 with an aim towards Single Instruction, Multiple Data (SIMD) optimization. This compilation strategy is selected for its capacity to handle diverse targets, thereby extending our computational efficiency analysis. In the comparison stage, our meticulous examination includes different architectures, dissecting computational efficiency, resource utilization, and potential bottlenecks. This comprehensive examination unfolds novel findings, shedding light on the workings of sparse neural network optimizations and informing future endeavors in computation techniques for sparse matrices. These insights serve as valuable guideposts for future advancements in hardware-accelerated AI and high-performance computing.

\end{itemize}

\subsubsection{Inference Acceleration Experiments} 

\begin{table}[b]
\centering
\begin{tabular}{c c c c c c}
\toprule
\multirow{2}{*}{Model} & \multicolumn{5}{c}{ \makecell{Inference time on different frameworks \\ mean (ms) / std (ms)}} \\  \cline{2-6}
\addlinespace[2pt]
 & PyTorch & TensorFlow & \small \makecell{LLVM generic\\dense} & \small \makecell{LLVM generic\\ sparsity 32 * 1} & \small 
 \makecell{LLVM AVX2 \\sparsity 64 * 1} \\ [8pt]\hline
\addlinespace[3pt]
BERT Base & 48.01 / 1.95 & 88.91 / 2.42 & 74.42 / 2.07 & 16.19 / 0.11 & \textbf{14.47 / 0.29}  \\ [2pt]\hline
\addlinespace[3pt]
BERT Large & 160.48 /  7.15 & 300.08 / 25.55 & 340.74 / 0.23 &54.82 / 0.25 & \textbf{47.64 / 0.37} \\ [2pt]\bottomrule \\
\end{tabular}
\caption{Comparative Performance Evaluations of Inference Engines. This table showcases the results of using Tensorflow, PyTorch, and LLVM compile, the latter with an intermediate Relay representation, as inference engines. Performance metrics were recorded across different settings with BERT Base and BERT Large models, running on an AMD Ryzen 9 5900X CPU. Each experiment was run five times, and the average mean and standard deviation (std) of the results are presented.}
\end{table}

To ascertain comparative baselines, initial evaluations were undertaken, deploying standard dense computations. These computations were facilitated by Tensorflow \citep{abadi2016tensorflow}, PyTorch \citep{paszke2019pytorch}, and LLVM compile \citep{chen2018tvm}, utilized as inference engines. For all LLVM compiling settings, an intermediate representation, Relay, was adopted, followed by execution of the Relay computation graph. This process is thoroughly elaborated in Figure \ref{fig:overview}.

Table 1 presents the performance outcomes under various settings, each experiment conducted five times to ensure robustness. Subsequently, the mean and standard deviation (std) of the results were reported to provide an accurate and consistent measure of performance. The inference experiments employed BERT Base and BERT Large models, run on a CPU - the AMD Ryzen 9 5900X processor.

Subsequently, the performance of these baselines was compared with that of sparse model variants, specifically irregular and block sparsity, using the SparseOptimizer applied to the BERT model. The best performance for LLVM generic compile, utilizing our co-design optimization, was found with sparsity block shapes of $32 * 1$, and for LLVM AVX2 compile, the optimal performance was achieved with block shapes of $64 * 1$.

Our Optimizer-Compiler Co-Design approach yielded significant acceleration of inference time for both BERT Base and BERT Large models. For the BERT Base model, the inference time was accelerated by a factor of \textbf{3.32x}, \textbf{6.14x}, and \textbf{5.14x} in comparison with Pytorch, TensorFlow, and LLVM generic compile, respectively. As for the BERT Large model, acceleration factors of \textbf{3.37x}, \textbf{6.30x}, and \textbf{7.15x} were observed, respectively.

\subsubsection{Compare Sparsity Block Shape}

Block sparsity is key to accelerating deep neural network inference time due to its alignment with hardware efficiency \citep{han2016eie, gray2017gpu, elsen2020fast, guo2021algorithm}. The shape of block sparsity, whether 1-D or 2-D, significantly influences computational efficiency and model performance \citep{wen2016learning, narang2017block, xu2018structured}.

A comparison of different block sparsity shapes helps identify the optimal trade-off between computational efficiency and model performance \citep{gale2020sparse}. This is particularly relevant for transformer-based models like BERT, where certain coarse-grained block sparsity shapes can deliver performance similar to dense models with substantial acceleration \citep{tay2020sparse}. Hence, in the context of our optimizer-compiler co-design, a detailed comparison will be pivotal in determining the most efficient block sparsity shape.

\begin{figure}[t]
    \centering
    \begin{subfigure}[b]{0.49\textwidth}
        \centering
        \includegraphics[width=\textwidth]{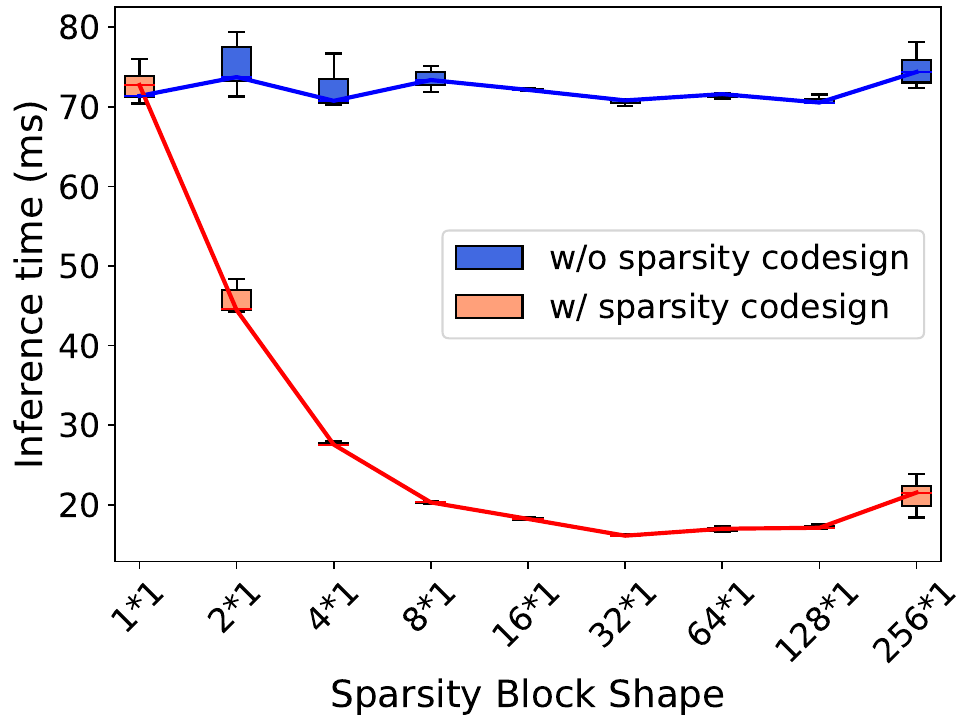}
        \caption{LLVM generic compile BERT Base}
        \label{fig:sub1_large}
    \end{subfigure}
    \hfill
    \begin{subfigure}[b]{0.49\textwidth}
        \centering
        \includegraphics[width=\textwidth]{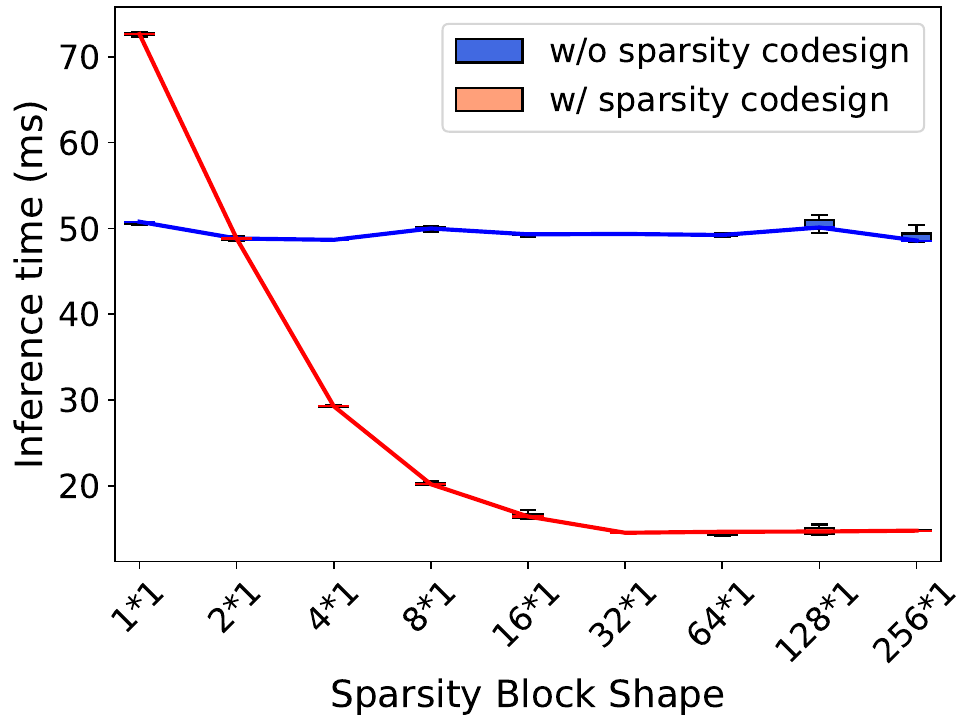}
        \caption{LLVM AVX2 compile BERT Base}
        \label{fig:sub2_large}
    \end{subfigure}
    
    \vspace*{5pt}
    
    \begin{subfigure}[b]{0.49\textwidth}
        \centering
        \includegraphics[width=\textwidth]{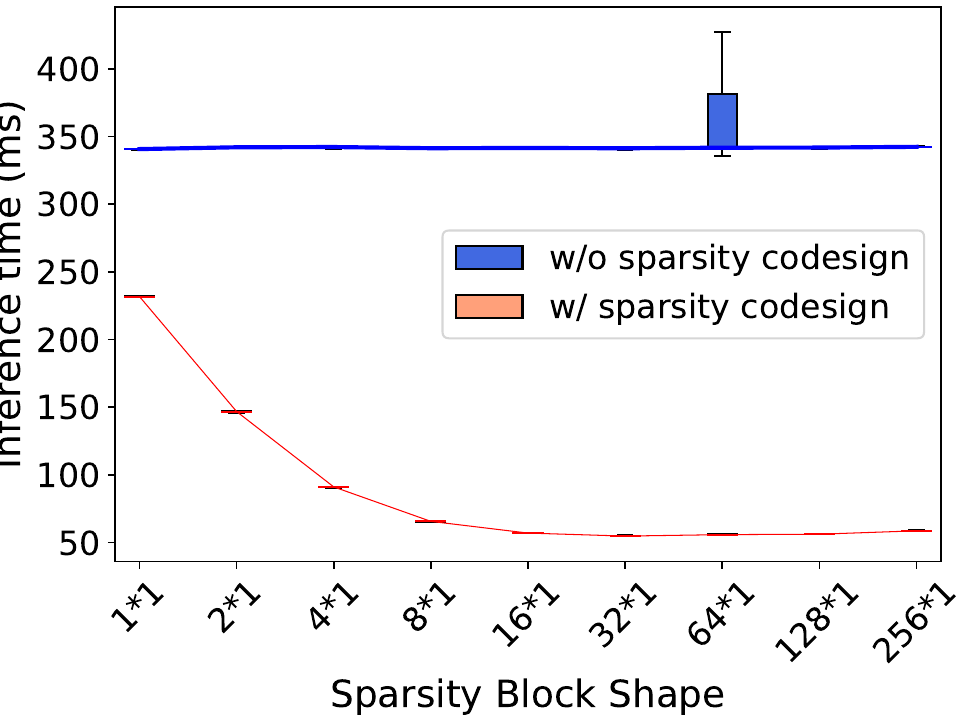}
        \caption{LLVM generic compile BERT Large}
        \label{fig:sub1}
    \end{subfigure}
    \hfill
    \begin{subfigure}[b]{0.49\textwidth}
        \centering
        \includegraphics[width=\textwidth]{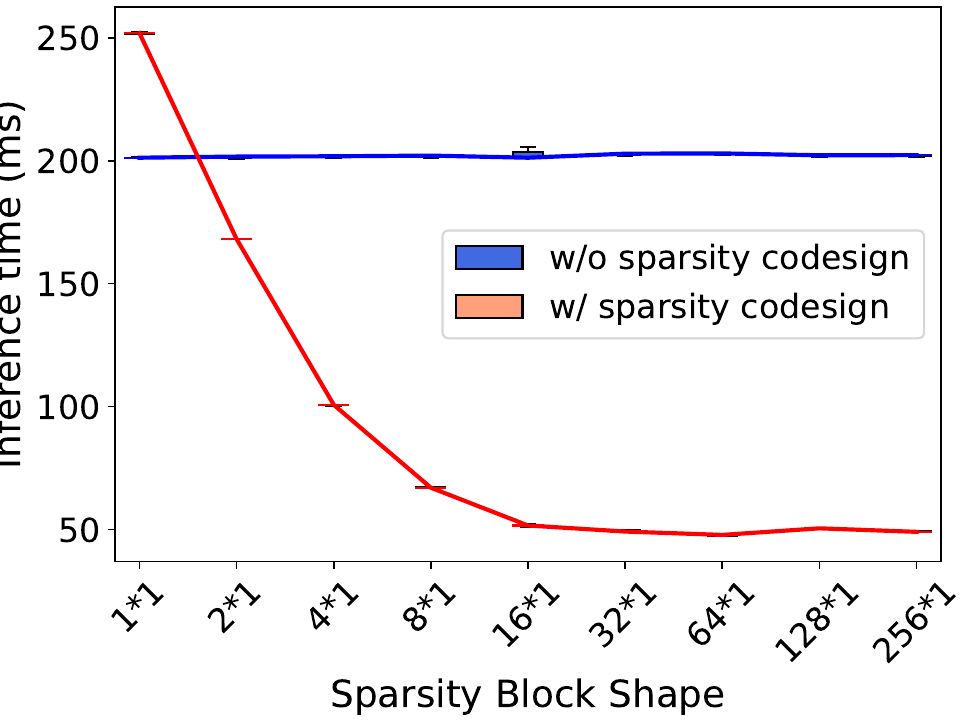}
        \caption{LLVM AVX2 compile BERT Large}
        \label{fig:sub2}
    \end{subfigure}
    
    \caption{Inference performance benchmark for SparseOptimizer-Compiler co-design. The intermediate representation of the computation graph is Relay \citep{roesch2018relay}, and the compiler is modified TVM\citep{chen2018tvm}. \textbf{Note:} both the blue line and the red line represent the inference time of the computing graph after compilation. The difference is that the blue line represents compiling the computation graph without explicitly analysing and optimization for the sparsity of the model, and the red line represents compiling the computation graph with explicitly analysing and optimization for the sparsity of the model. Sub-figure (a) and (c) uses the LLVM generic compilation and the sub-figure (b) and (d) uses the LLVM AVX2 compilation with SIMD optimization. All the results are on AMD Ryzen 9 5900X processor. }
    \label{fig:box}
\end{figure}

In this study, we examine the influence of block sparsity shape on inference time through a series of controlled experiments. We leverage two compiler configurations: the generic LLVM compiler aimed at a broad CPU architecture and the LLVM AVX 2 compiler equipped with SIMD optimization, explicitly targeting CPUs supporting AVX 2, including the Intel Haswell family and AMD Zen family. Specifically, all experiments are conducted on an AMD Ryzen 9 5900X processor. The experimental framework includes both BERT Base and BERT Large models. In each environment, we implement the linear block shape "n*1", $(n\in {1, 2, 4, 8, 16, 32, 64, 128, 256})$, incorporating sparsity compiler co-design. For each block size within each setting, the model undergoes inference five times. The results, which include maximum, minimum, median, mean, and standard deviation, are represented via a Box and Whisker Plot (refer to Figure \ref{fig:box}). The median values are connected to form a trend line in the figure, which provides a robust visualization as it is less susceptible to outliers compared to the mean.

We have the following observations through the Figure \ref{fig:box}:
\pagebreak

\begin{itemize}
    \item Across both LLVM generic and AVX2 compilation configurations, as well as BERT Base and Large models, compilation employing sparsity co-design (denoted by the red line in Figure \ref{fig:box}) invariably exhibits superior performance relative to compilation devoid of sparsity co-design (represented by the blue line in Figure \ref{fig:box}).
    
    \item The relationship between linear block size and computation duration exhibits a non-monotonic behavior. Specifically, there is a notable reduction in inference time as block sparsity dimensions transition from $1 * 1$ to $32 * 1$ ($64 * 1$ for LLVM AVX2 compile). However, an increase in the sparsity dimensions beyond this range results in an elevated computation time, demonstrating a deterioration in performance.
    
    \item The deterioration after the optimal point is not that obvious for LLVM AVX2 compiling with SIMD optimization, compared to LLVM generic compiling (Figure \ref{fig:sub1}).

    \item The ideal block sparsity shape for LLVM AVX2 compilation is $64 * 1$ (Figure \ref{fig:sub2} and \ref{fig:sub2_large}), which is larger than the optimal shape of $32 * 1$ (Figure \ref{fig:sub1} and \ref{fig:sub1_large}) for LLVM generic compilation. This difference could be attributed to the SIMD optimization.

    \item There are evident intersections between the red and blue lines in Figure \ref{fig:sub2} and \ref{fig:sub2_large}, contrasting with Figure \ref{fig:sub1} and \ref{fig:sub1_large}. This indicates that LLVM AVX2 compilation can directly accelerate sparsity for sparse BERT models, even in the absence of sparsity co-design. Nevertheless, as we expand the block size from $1*1$ to $64*1$ and then $256*1$, the performance of the sparsity co-design method surpasses compilation without sparsity co-design.

\end{itemize}

\section{Conclusion and Discussion}

This research paper makes significant contributions to the development and efficient deployment of Large Language Models (LLMs) by introducing the innovative SparseOptimizer. This unique deep learning optimizer, designed specifically for LLMs, is characterized by its seamless plug-and-play design, enabling its direct integration into LLMs without necessitating any modifications to the original model code.

The SparseOptimizer is particularly effective in inducing sparsity in model parameters, enhancing the efficiency of LLMs. We demonstrated the efficacy of SparseBERT and SparseALBERT models across several NLP benchmarks such as GLUE\citep{wang2018glue}, SQuAD v1.1\citep{rajpurkar2016squad}, SQuAD v2.0\citep{rajpurkar2018know}, and RACE\citep{lai2017race}. Despite significant reductions in their model sizes, these models displayed impressive efficiency. For instance, the model sizes for SparseALBERT Base and SparseALBERT Large were brought down to a mere \textbf{5M} and 8M parameters, respectively, without sacrificing performance. This size efficiency renders these models highly desirable for deployment in resource-constrained environments.

Furthermore, our work reveals an innovative optimizer-compiler co-design strategy that has substantially improved the efficient execution of sparse models. The strategy's effectiveness is illustrated by substantial inference acceleration, with improvements of \textbf{3.37x}, \textbf{6.30x}, and \textbf{7.15x} over Pytorch, TensorFlow, and LLVM generic compile, respectively. These performance enhancements were achieved when we applied our co-design strategy in combination with an appropriately designed compiler for SparseBERT models.

However, we also observed a non-monotonic relationship between block size and computation duration. Interestingly, this performance deterioration beyond the optimal block sparsity dimensions can be mitigated by the SIMD optimization of LLVM AVX2 compilation, which can also accelerate sparsity for sparse models without the need for co-design.

In conclusion, our work represents a significant leap forward in the practical and efficient deployment of LLMs. By explicitly highlighting the plug-and-play nature of SparseOptimizer, we underscore its broad applicability and ease of use in a wide variety of contexts. The successful implementation and performance of SparseOptimizer, coupled with our optimizer-compiler co-design strategy, pave the way for future explorations in harnessing the power of LLMs in resource-limited settings while maintaining high levels of performance and efficiency.

\begin{ack}
The authors gratefully acknowledge the support of the computing resources sponsor: Google TPU Research Cloud and Google LLC. Both the authors and this work are affiliated with Lotus AGI, Corp.
\end{ack}

\medskip

\bibliographystyle{abbrvnat}
\bibliography{neurips_2023}

\begin{thebibliography}{72}
\providecommand{\natexlab}[1]{#1}
\providecommand{\url}[1]{\texttt{#1}}
\expandafter\ifx\csname urlstyle\endcsname\relax
  \providecommand{\doi}[1]{doi: #1}\else
  \providecommand{\doi}{doi: \begingroup \urlstyle{rm}\Url}\fi

\bibitem[Abadi et~al.(2016)Abadi, Barham, Chen, Chen, Davis, Dean, Devin,
  Ghemawat, Irving, et~al.]{abadi2016tensorflow}
M.~A. Abadi, P.~Barham, J.~Chen, Z.~Chen, A.~Davis, J.~Dean, M.~Devin,
  S.~Ghemawat, G.~Irving, et~al.
\newblock Tensorflow: A system for large-scale machine learning.
\newblock In \emph{12th USENIX Symposium on Operating Systems Design and
  Implementation (OSDI 16)}, pages 265--283, 2016.

\bibitem[Agirre and Soroa(2007)]{agirre2007semeval}
E.~Agirre and A.~Soroa.
\newblock Semeval-2007 task 02: Evaluating word sense induction and
  discrimination systems.
\newblock In \emph{Proceedings of the 4th International Workshop on Semantic
  Evaluations}, pages 7--12. Association for Computational Linguistics, 2007.

\bibitem[Bacho(2023)]{bacho2023generalization}
A.~Bacho.
\newblock A generalization of the moreau--yosida regularization.
\newblock \emph{Journal of Mathematical Analysis and Applications},
  524\penalty0 (2):\penalty0 127139, 2023.

\bibitem[Bahdanau et~al.(2014)Bahdanau, Cho, and Bengio]{bahdanau2014neural}
D.~Bahdanau, K.~Cho, and Y.~Bengio.
\newblock Neural machine translation by jointly learning to align and
  translate.
\newblock \emph{arXiv preprint arXiv:1409.0473}, 2014.

\bibitem[Boyd et~al.(2011)Boyd, Parikh, Chu, Peleato, Eckstein,
  et~al.]{boyd2011distributed}
S.~Boyd, N.~Parikh, E.~Chu, B.~Peleato, J.~Eckstein, et~al.
\newblock Distributed optimization and statistical learning via the alternating
  direction method of multipliers.
\newblock \emph{Foundations and Trends{\textregistered} in Machine learning},
  3\penalty0 (1):\penalty0 1--122, 2011.

\bibitem[Boyd and Vandenberghe(2004)]{boyd2004convex}
S.~P. Boyd and L.~Vandenberghe.
\newblock \emph{Convex optimization}.
\newblock Cambridge university press, 2004.

\bibitem[Brown et~al.(2020)Brown, Mann, Ryder, Subbiah, Kaplan, Dhariwal,
  Neelakantan, Shyam, Sastry, Askell, et~al.]{brown2020language}
T.~Brown, B.~Mann, N.~Ryder, M.~Subbiah, J.~D. Kaplan, P.~Dhariwal,
  A.~Neelakantan, P.~Shyam, G.~Sastry, A.~Askell, et~al.
\newblock Language models are few-shot learners.
\newblock \emph{Advances in neural information processing systems},
  33:\penalty0 1877--1901, 2020.

\bibitem[Candes et~al.(2008)Candes, Wakin, and Boyd]{candes2008enhancing}
E.~J. Candes, M.~B. Wakin, and S.~P. Boyd.
\newblock Enhancing sparsity by reweighted $\ell_1$ minimization.
\newblock \emph{Journal of Fourier analysis and applications}, 14:\penalty0
  877--905, 2008.

\bibitem[Chen et~al.(2018)Chen, Moreau, Jiang, Zheng, Yan, Cowan, Shen, Wang,
  Hu, Ceze, et~al.]{chen2018tvm}
T.~Chen, T.~Moreau, Z.~Jiang, L.~Zheng, E.~Yan, M.~Cowan, H.~Shen, L.~Wang,
  Y.~Hu, L.~Ceze, et~al.
\newblock Tvm: an automated end-to-end optimizing compiler for deep learning.
\newblock In \emph{Proceedings of the 13th USENIX conference on Operating
  Systems Design and Implementation}, pages 579--594, 2018.

\bibitem[Chen et~al.(2020)Chen, Frankle, Chang, Liu, Zhang, Wang, and
  Carbin]{chen2020lottery}
T.~Chen, J.~Frankle, S.~Chang, S.~Liu, Y.~Zhang, Z.~Wang, and M.~Carbin.
\newblock The lottery ticket hypothesis for pre-trained bert networks.
\newblock \emph{Advances in neural information processing systems},
  33:\penalty0 15834--15846, 2020.

\bibitem[Chen et~al.(2021)Chen, Frankle, Chang, Liu, Zhang, Carbin, and
  Wang]{chen2021lottery}
T.~Chen, J.~Frankle, S.~Chang, S.~Liu, Y.~Zhang, M.~Carbin, and Z.~Wang.
\newblock The lottery tickets hypothesis for supervised and self-supervised
  pre-training in computer vision models.
\newblock \emph{arXiv preprint arXiv:2012.06908}, 2021.

\bibitem[Dettmers et~al.(2023)Dettmers, Pagnoni, Holtzman, and
  Zettlemoyer]{dettmers2023qlora}
T.~Dettmers, A.~Pagnoni, A.~Holtzman, and L.~Zettlemoyer.
\newblock Qlora: Efficient finetuning of quantized llms.
\newblock \emph{arXiv preprint arXiv:2305.14314}, 2023.

\bibitem[Devlin et~al.(2018)Devlin, Chang, Lee, and Toutanova]{devlin2018bert}
J.~Devlin, M.-W. Chang, K.~Lee, and K.~Toutanova.
\newblock Bert: Pre-training of deep bidirectional transformers for language
  understanding.
\newblock \emph{arXiv preprint arXiv:1810.04805}, 2018.

\bibitem[Devlin et~al.(2019)Devlin, Chang, Lee, and Toutanova]{devlin2019bert}
J.~Devlin, M.-W. Chang, K.~Lee, and K.~Toutanova.
\newblock Bert: Pre-training of deep bidirectional transformers for language
  understanding.
\newblock In \emph{Proceedings of the 2019 Conference of the North American
  Chapter of the Association for Computational Linguistics: Human Language
  Technologies, Volume 1 (Long and Short Papers)}, pages 4171--4186, 2019.

\bibitem[Dolan and Brockett(2005)]{dolan2005automatically}
W.~B. Dolan and C.~Brockett.
\newblock Automatically constructing a corpus of sentential paraphrases.
\newblock In \emph{Proceedings of the Third International Workshop on
  Paraphrasing (IWP2005)}, 2005.

\bibitem[Elsen et~al.(2020)Elsen, Dukhan, Gale, and Simonyan]{elsen2020fast}
E.~Elsen, M.~Dukhan, T.~Gale, and K.~Simonyan.
\newblock Fast sparse convnets.
\newblock In \emph{Proceedings of the IEEE/CVF conference on computer vision
  and pattern recognition}, pages 14629--14638, 2020.

\bibitem[Frankle and Carbin(2018)]{frankle2018lottery}
J.~Frankle and M.~Carbin.
\newblock The lottery ticket hypothesis: Finding sparse, trainable neural
  networks.
\newblock \emph{arXiv preprint arXiv:1803.03635}, 2018.

\bibitem[Frankle and Carbin(2019)]{franklelottery}
J.~Frankle and M.~Carbin.
\newblock The lottery ticket hypothesis: Finding sparse, trainable neural
  networks.
\newblock In \emph{International Conference on Learning Representations}, 2019.

\bibitem[Frankle et~al.(2020)Frankle, Dziugaite, Roy, and
  Carbin]{frankle2020linear}
J.~Frankle, G.~K. Dziugaite, D.~Roy, and M.~Carbin.
\newblock Linear mode connectivity and the lottery ticket hypothesis.
\newblock In \emph{International Conference on Machine Learning}, pages
  3259--3269. PMLR, 2020.

\bibitem[Gale et~al.(2020)Gale, Zaharia, Young, and Elsen]{gale2020sparse}
T.~Gale, M.~Zaharia, C.~Young, and E.~Elsen.
\newblock Sparse gpu kernels for deep learning.
\newblock In \emph{SC20: International Conference for High Performance
  Computing, Networking, Storage and Analysis}, pages 1--14. IEEE, 2020.

\bibitem[Google(2023)]{anil2023palm}
Google.
\newblock Palm 2 technical report.
\newblock \emph{arXiv preprint arXiv:2305.10403}, 2023.

\bibitem[Gray et~al.(2017)Gray, Radford, and Kingma]{gray2017gpu}
S.~Gray, A.~Radford, and D.~P. Kingma.
\newblock Gpu kernels for block-sparse weights.
\newblock \emph{arXiv preprint arXiv:1711.09224}, 3, 2017.

\bibitem[Guo and Huang(2021)]{guo2021algorithm}
F.-M. Guo and A.~Huang.
\newblock Algorithm to compilation co-design: An integrated view of neural
  network sparsity.
\newblock \emph{arXiv preprint arXiv:2106.08846}, 2021.

\bibitem[Guo et~al.(2019)Guo, Liu, Mungall, Lin, and Wang]{guo2019reweighted}
F.-M. Guo, S.~Liu, F.~S. Mungall, X.~Lin, and Y.~Wang.
\newblock Reweighted proximal pruning for large-scale language representation.
\newblock \emph{arXiv preprint arXiv:1909.12486}, 2019.

\bibitem[Han et~al.(2015)Han, Mao, and Dally]{han2015deep}
S.~Han, H.~Mao, and W.~J. Dally.
\newblock Deep compression: Compressing deep neural networks with pruning,
  trained quantization and huffman coding.
\newblock \emph{arXiv preprint arXiv:1510.00149}, 2015.

\bibitem[Han et~al.(2016)Han, Liu, Mao, Pu, Pedram, Horowitz, and
  Dally]{han2016eie}
S.~Han, X.~Liu, H.~Mao, J.~Pu, A.~Pedram, M.~A. Horowitz, and W.~J. Dally.
\newblock Eie: Efficient inference engine on compressed deep neural network.
\newblock \emph{ACM SIGARCH Computer Architecture News}, 44\penalty0
  (3):\penalty0 243--254, 2016.

\bibitem[Hassibi and Stork(1992)]{hassibi1992second}
B.~Hassibi and D.~Stork.
\newblock Second order derivatives for network pruning: Optimal brain surgeon.
\newblock \emph{Advances in neural information processing systems}, 5, 1992.

\bibitem[Hinton et~al.(2015)Hinton, Vinyals, and Dean]{hinton2015distilling}
G.~Hinton, O.~Vinyals, and J.~Dean.
\newblock Distilling the knowledge in a neural network.
\newblock \emph{arXiv preprint arXiv:1503.02531}, 2015.

\bibitem[Hsieh et~al.(2023)Hsieh, Li, Yeh, Nakhost, Fujii, Ratner, Krishna,
  Lee, and Pfister]{hsieh2023distilling}
C.-Y. Hsieh, C.-L. Li, C.-K. Yeh, H.~Nakhost, Y.~Fujii, A.~Ratner, R.~Krishna,
  C.-Y. Lee, and T.~Pfister.
\newblock Distilling step-by-step! outperforming larger language models with
  less training data and smaller model sizes.
\newblock \emph{arXiv preprint arXiv:2305.02301}, 2023.

\bibitem[Jiao et~al.(2019)Jiao, Yin, Shang, Jiang, Chen, Li, Wang, and
  Liu]{jiao2019tinybert}
X.~Jiao, Y.~Yin, L.~Shang, X.~Jiang, X.~Chen, L.~Li, F.~Wang, and Q.~Liu.
\newblock Tinybert: Distilling bert for natural language understanding.
\newblock \emph{arXiv preprint arXiv:1909.10351}, 2019.

\bibitem[Kim et~al.(2021)Kim, Gholami, Yao, Mahoney, and Keutzer]{kim2021bert}
S.~Kim, A.~Gholami, Z.~Yao, M.~W. Mahoney, and K.~Keutzer.
\newblock I-bert: Integer-only bert quantization.
\newblock In \emph{International conference on machine learning}, pages
  5506--5518. PMLR, 2021.

\bibitem[Lai et~al.(2017)Lai, Xie, Liu, Yang, and Hovy]{lai2017race}
G.~Lai, Q.~Xie, H.~Liu, Y.~Yang, and E.~Hovy.
\newblock Race: Large-scale reading comprehension dataset from examinations.
\newblock In \emph{Proceedings of the 2017 Conference on Empirical Methods in
  Natural Language Processing}, pages 785--794, 2017.

\bibitem[Lan et~al.(2019)Lan, Chen, Goodman, Gimpel, Sharma, and
  Soricut]{lan2019albert}
Z.~Lan, M.~Chen, S.~Goodman, K.~Gimpel, P.~Sharma, and R.~Soricut.
\newblock Albert: A lite bert for self-supervised learning of language
  representations.
\newblock \emph{arXiv preprint arXiv:1909.11942}, 2019.

\bibitem[LeCun et~al.(1989)LeCun, Denker, and Solla]{lecun1989optimal}
Y.~LeCun, J.~Denker, and S.~Solla.
\newblock Optimal brain damage.
\newblock \emph{Advances in neural information processing systems}, 2, 1989.

\bibitem[Levesque et~al.(2012)Levesque, Davis, and
  Morgenstern]{levesque2012winograd}
H.~Levesque, E.~Davis, and L.~Morgenstern.
\newblock The winograd schema challenge.
\newblock In \emph{Thirteenth International Conference on the Principles of
  Knowledge Representation and Reasoning}, 2012.

\bibitem[Lin et~al.(2013)Lin, Fardad, and Jovanovi{\'c}]{lin2013design}
F.~Lin, M.~Fardad, and M.~R. Jovanovi{\'c}.
\newblock Design of optimal sparse feedback gains via the alternating direction
  method of multipliers.
\newblock \emph{IEEE Transactions on Automatic Control}, 58\penalty0
  (9):\penalty0 2426--2431, 2013.

\bibitem[Liu et~al.(2014)Liu, Masazade, Fardad, and Varshney]{liu2014sparsity}
S.~Liu, E.~Masazade, M.~Fardad, and P.~K. Varshney.
\newblock Sparsity-aware field estimation via ordinary kriging.
\newblock In \emph{2014 IEEE International Conference on Acoustics, Speech and
  Signal Processing (ICASSP)}, pages 3948--3952. IEEE, 2014.

\bibitem[Loshchilov and Hutter(2017)]{loshchilov2017decoupled}
I.~Loshchilov and F.~Hutter.
\newblock Decoupled weight decay regularization.
\newblock \emph{arXiv preprint arXiv:1711.05101}, 2017.

\bibitem[Martinet(1970)]{martinet1970breve}
B.~Martinet.
\newblock Br{\`e}ve communication. r{\'e}gularisation d'in{\'e}quations
  variationnelles par approximations successives.
\newblock \emph{Revue fran{\c{c}}aise d'informatique et de recherche
  op{\'e}rationnelle. S{\'e}rie rouge}, 4\penalty0 (R3):\penalty0 154--158,
  1970.

\bibitem[Michel et~al.(2019)Michel, Levy, and Neubig]{michel2019sixteen}
P.~Michel, O.~Levy, and G.~Neubig.
\newblock Are sixteen heads really better than one?
\newblock \emph{Advances in neural information processing systems}, 32, 2019.

\bibitem[Moreau(1966)]{moreau1966fonctionnelles}
J.-J. Moreau.
\newblock Fonctionnelles convexes.
\newblock \emph{S{\'e}minaire Jean Leray}, 1\penalty0 (2):\penalty0 1--108,
  1966.

\bibitem[Narang et~al.(2017)Narang, Undersander, and Diamos]{narang2017block}
S.~Narang, E.~Undersander, and G.~Diamos.
\newblock Block-sparse recurrent neural networks.
\newblock \emph{arXiv preprint arXiv:1711.02782}, 2017.

\bibitem[OpenAI(2023)]{gpt4}
OpenAI.
\newblock Gpt4: Technical report.
\newblock \emph{arXiv preprint arXiv:2303.08774}, 2023.

\bibitem[Ouyang et~al.(2022)Ouyang, Wu, Jiang, Almeida, Wainwright, Mishkin,
  Zhang, Agarwal, Slama, Ray, et~al.]{ouyang2022training}
L.~Ouyang, J.~Wu, X.~Jiang, D.~Almeida, C.~Wainwright, P.~Mishkin, C.~Zhang,
  S.~Agarwal, K.~Slama, A.~Ray, et~al.
\newblock Training language models to follow instructions with human feedback.
\newblock \emph{Advances in Neural Information Processing Systems},
  35:\penalty0 27730--27744, 2022.

\bibitem[Parikh et~al.(2014)Parikh, Boyd, et~al.]{parikh2014proximal}
N.~Parikh, S.~Boyd, et~al.
\newblock Proximal algorithms.
\newblock \emph{Foundations and trends{\textregistered} in Optimization},
  1\penalty0 (3):\penalty0 127--239, 2014.

\bibitem[Paszke et~al.(2019)Paszke, Gross, Massa, Lerer, Bradbury, Chanan,
  Killeen, Lin, Gimelshein, Antiga, et~al.]{paszke2019pytorch}
A.~Paszke, S.~Gross, F.~Massa, A.~Lerer, J.~Bradbury, G.~Chanan, T.~Killeen,
  Z.~Lin, N.~Gimelshein, L.~Antiga, et~al.
\newblock Pytorch: An imperative style, high-performance deep learning library.
\newblock \emph{Advances in neural information processing systems}, 32, 2019.

\bibitem[Patterson et~al.(2021)Patterson, Gonzalez, Le, Liang, Munguia,
  Rothchild, So, Texier, and Dean]{patterson2021carbon}
D.~Patterson, J.~Gonzalez, Q.~Le, C.~Liang, L.-M. Munguia, D.~Rothchild, D.~So,
  M.~Texier, and J.~Dean.
\newblock Carbon emissions and large neural network training.
\newblock \emph{arXiv preprint arXiv:2104.10350}, 2021.

\bibitem[Rajpurkar et~al.(2016)Rajpurkar, Zhang, Lopyrev, and
  Liang]{rajpurkar2016squad}
P.~Rajpurkar, J.~Zhang, K.~Lopyrev, and P.~Liang.
\newblock Squad: 100,000+ questions for machine comprehension of text.
\newblock \emph{arXiv preprint arXiv:1606.05250}, 2016.

\bibitem[Rajpurkar et~al.(2018)Rajpurkar, Jia, and Liang]{rajpurkar2018know}
P.~Rajpurkar, R.~Jia, and P.~Liang.
\newblock Know what you don't know: Unanswerable questions for squad.
\newblock \emph{arXiv preprint arXiv:1806.03822}, 2018.

\bibitem[Roesch et~al.(2018)Roesch, Lyubomirsky, Weber, Pollock, Kirisame,
  Chen, and Tatlock]{roesch2018relay}
J.~Roesch, S.~Lyubomirsky, L.~Weber, J.~Pollock, M.~Kirisame, T.~Chen, and
  Z.~Tatlock.
\newblock Relay: A new ir for machine learning frameworks.
\newblock In \emph{Proceedings of the 2nd ACM SIGPLAN international workshop on
  machine learning and programming languages}, pages 58--68, 2018.

\bibitem[Rosenbrock(1960)]{rosenbrock1960automatic}
H.~Rosenbrock.
\newblock An automatic method for finding the greatest or least value of a
  function.
\newblock \emph{The computer journal}, 3\penalty0 (3):\penalty0 175--184, 1960.

\bibitem[Sanh et~al.(2019)Sanh, Debut, Chaumond, and Wolf]{sanh2019distilbert}
V.~Sanh, L.~Debut, J.~Chaumond, and T.~Wolf.
\newblock Distilbert, a distilled version of bert: smaller, faster, cheaper and
  lighter.
\newblock \emph{arXiv preprint arXiv:1910.01108}, 2019.

\bibitem[Sanh et~al.(2020)Sanh, Wolf, and Rush]{sanh2020movement}
V.~Sanh, T.~Wolf, and A.~Rush.
\newblock Movement pruning: Adaptive sparsity by fine-tuning.
\newblock \emph{Advances in Neural Information Processing Systems},
  33:\penalty0 20378--20389, 2020.

\bibitem[Socher et~al.(2013)Socher, Perelygin, Wu, Chuang, Manning, Ng, and
  Potts]{socher2013recursive}
R.~Socher, A.~Perelygin, J.~Wu, J.~Chuang, C.~D. Manning, A.~Ng, and C.~Potts.
\newblock Recursive deep models for semantic compositionality over a sentiment
  treebank.
\newblock In \emph{Proceedings of the 2013 conference on empirical methods in
  natural language processing}, pages 1631--1642, 2013.

\bibitem[Str{\"o}m(1997)]{strom1997phoneme}
N.~Str{\"o}m.
\newblock Phoneme probability estimation with dynamic sparsely connected
  artificial neural networks.
\newblock \emph{The Free Speech Journal}, 5\penalty0 (1-41):\penalty0 2, 1997.

\bibitem[Sun et~al.(2020)Sun, Yu, Song, Liu, Yang, and Zhou]{sun2020mobilebert}
Z.~Sun, H.~Yu, X.~Song, R.~Liu, Y.~Yang, and D.~Zhou.
\newblock Mobilebert: a compact task-agnostic bert for resource-limited
  devices.
\newblock \emph{arXiv preprint arXiv:2004.02984}, 2020.

\bibitem[Tay et~al.(2020)Tay, Bahri, Yang, Metzler, and Juan]{tay2020sparse}
Y.~Tay, D.~Bahri, L.~Yang, D.~Metzler, and D.-C. Juan.
\newblock Sparse sinkhorn attention.
\newblock In \emph{International Conference on Machine Learning}, pages
  9438--9447. PMLR, 2020.

\bibitem[Thoppilan et~al.(2022)Thoppilan, De~Freitas, Hall, Shazeer,
  Kulshreshtha, Cheng, Jin, Bos, Baker, Du, et~al.]{thoppilan2022lamda}
R.~Thoppilan, D.~De~Freitas, J.~Hall, N.~Shazeer, A.~Kulshreshtha, H.-T. Cheng,
  A.~Jin, T.~Bos, L.~Baker, Y.~Du, et~al.
\newblock Lamda: Language models for dialog applications.
\newblock \emph{arXiv preprint arXiv:2201.08239}, 2022.

\bibitem[Turc et~al.(2019)Turc, Chang, Lee, and Toutanova]{turc2019well}
I.~Turc, M.-W. Chang, K.~Lee, and K.~Toutanova.
\newblock Well-read students learn better: On the importance of pre-training
  compact models.
\newblock \emph{arXiv preprint arXiv:1908.08962}, 2019.

\bibitem[Vaswani et~al.(2017)Vaswani, Shazeer, Parmar, Uszkoreit, Jones, Gomez,
  Kaiser, and Polosukhin]{vaswani2017attention}
A.~Vaswani, N.~Shazeer, N.~Parmar, J.~Uszkoreit, L.~Jones, A.~N. Gomez,
  {\L}.~Kaiser, and I.~Polosukhin.
\newblock Attention is all you need.
\newblock \emph{Advances in neural information processing systems}, 30, 2017.

\bibitem[Voita et~al.(2019)Voita, Talbot, Moiseev, Sennrich, and
  Titov]{voita2019analyzing}
E.~Voita, D.~Talbot, F.~Moiseev, R.~Sennrich, and I.~Titov.
\newblock Analyzing multi-head self-attention: Specialized heads do the heavy
  lifting, the rest can be pruned.
\newblock \emph{arXiv preprint arXiv:1905.09418}, 2019.

\bibitem[Wang et~al.(2018)Wang, Singh, Michael, Hill, Levy, and
  Bowman]{wang2018glue}
A.~Wang, A.~Singh, J.~Michael, F.~Hill, O.~Levy, and S.~R. Bowman.
\newblock Glue: A multi-task benchmark and analysis platform for natural
  language understanding.
\newblock \emph{arXiv preprint arXiv:1804.07461}, 2018.

\bibitem[Wang et~al.(2020)Wang, Wei, Dong, Bao, Yang, and Zhou]{wang2020minilm}
W.~Wang, F.~Wei, L.~Dong, H.~Bao, N.~Yang, and M.~Zhou.
\newblock Minilm: Deep self-attention distillation for task-agnostic
  compression of pre-trained transformers.
\newblock \emph{Advances in Neural Information Processing Systems},
  33:\penalty0 5776--5788, 2020.

\bibitem[Warstadt et~al.(2018)Warstadt, Singh, and Bowman]{warstadt2018neural}
A.~Warstadt, A.~Singh, and S.~R. Bowman.
\newblock Neural network acceptability judgments.
\newblock \emph{arXiv preprint arXiv:1805.12471}, 2018.

\bibitem[Wen et~al.(2016)Wen, Wu, Wang, Chen, and Li]{wen2016learning}
W.~Wen, C.~Wu, Y.~Wang, Y.~Chen, and H.~Li.
\newblock Learning structured sparsity in deep neural networks.
\newblock \emph{Advances in neural information processing systems}, 29, 2016.

\bibitem[Williams et~al.(2017)Williams, Nangia, and Bowman]{williams2017broad}
A.~Williams, N.~Nangia, and S.~R. Bowman.
\newblock A broad-coverage challenge corpus for sentence understanding through
  inference.
\newblock \emph{arXiv preprint arXiv:1704.05426}, 2017.

\bibitem[Wu et~al.(2016)Wu, Schuster, Chen, Le, Norouzi, Macherey, Krikun, Cao,
  Gao, Macherey, et~al.]{wu2016google}
Y.~Wu, M.~Schuster, Z.~Chen, Q.~V. Le, M.~Norouzi, W.~Macherey, M.~Krikun,
  Y.~Cao, Q.~Gao, K.~Macherey, et~al.
\newblock Google's neural machine translation system: Bridging the gap between
  human and machine translation.
\newblock \emph{arXiv preprint arXiv:1609.08144}, 2016.

\bibitem[Xia et~al.(2022)Xia, Zhong, and Chen]{xia2022structured}
M.~Xia, Z.~Zhong, and D.~Chen.
\newblock Structured pruning learns compact and accurate models.
\newblock In \emph{Proceedings of the 60th Annual Meeting of the Association
  for Computational Linguistics (Volume 1: Long Papers)}, pages 1513--1528,
  2022.

\bibitem[Xu et~al.(2018)Xu, Liu, Zhao, Chen, Zhang, Fan, Erdogmus, Wang, and
  Lin]{xu2018structured}
K.~Xu, S.~Liu, P.~Zhao, P.-Y. Chen, H.~Zhang, Q.~Fan, D.~Erdogmus, Y.~Wang, and
  X.~Lin.
\newblock Structured adversarial attack: Towards general implementation and
  better interpretability.
\newblock \emph{arXiv preprint arXiv:1808.01664}, 2018.

\bibitem[Zhang et~al.(2021)Zhang, Huda, Songhori, Le, Goldie, and
  Mirhoseini]{zhang2021full}
D.~Zhang, S.~Huda, E.~Songhori, Q.~Le, A.~Goldie, and A.~Mirhoseini.
\newblock A full-stack accelerator search technique for vision applications.
\newblock \emph{arXiv preprint arXiv:2105.12842}, 2021.

\bibitem[Zhu and Gupta(2017)]{zhu2017prune}
M.~Zhu and S.~Gupta.
\newblock To prune, or not to prune: exploring the efficacy of pruning for
  model compression.
\newblock \emph{arXiv preprint arXiv:1710.01878}, 2017.

\bibitem[Zhu et~al.(2015)Zhu, Kiros, Zemel, Salakhutdinov, Urtasun, Torralba,
  and Fidler]{zhu2015aligning}
Y.~Zhu, R.~Kiros, R.~Zemel, R.~Salakhutdinov, R.~Urtasun, A.~Torralba, and
  S.~Fidler.
\newblock Aligning books and movies: Towards story-like visual explanations by
  watching movies and reading books.
\newblock In \emph{Proceedings of the IEEE international conference on computer
  vision}, pages 19--27, 2015.

\end{thebibliography}


\end{document}